\documentclass[11pt,a4paper]{article}
\usepackage[hyperref]{acl2021}
\usepackage{times}
\usepackage{latexsym}

\usepackage{graphicx}
\usepackage{multirow}
\usepackage{array,booktabs}
\usepackage{textcomp}
\usepackage{xcolor,colortbl}
\usepackage{amssymb}% http://ctan.org/pkg/amssymb
\usepackage{pifont}% http://ctan.org/pkg/pifont
\newcommand*\rot{\rotatebox{90}}

\usepackage{amsmath}
\usepackage{array, multirow, bigdelim, makecell, booktabs} 
\usepackage[inline]{enumitem}
\usepackage{xcolor}
\newcommand{\ensuretext}[1]{#1}
\newcommand{\jtmarker}{\ensuretext{\textcolor{blue}{\ensuremath{^{\textsc{J}}_{\textsc{T}}}}}}
\newcommand{\ebmarker}{\ensuretext{\textcolor{orange}{\ensuremath{^{\textsc{E}}_{\textsc{B}}}}}}
\newcommand{\mycomment}[3]{}

\newcommand{\eb}[1]{\mycomment{\ebmarker}{#1}{orange}}

\newcommand{\jt}[1]{\mycomment{\jtmarker}{#1}{blue}}

\newcommand{\ignore}[1]{}

\usepackage{microtype}

\aclfinalcopy % Uncomment this line for the final submission
\title{A Review of Human Evaluation for Style Transfer}

\author{Eleftheria Briakou$^1$ Sweta Agrawal$^1$ Ke Zhang$^2$ Joel Tetreault$^2$ Marine Carpuat$^1$  \\
  $^1$University of Maryland,
  $^2$Dataminr, Inc. \\
 \texttt{\href{mailto:ebriakou@cs.umd.edu}{ebriakou@cs.umd.edu}},
 \texttt{\href{mailto:sweagraw@cs.umd.edu}{sweagraw@cs.umd.edu}},
 \texttt{\href{mailto:kzhang@dataminr.com}{kzhang@dataminr.com}},\\
 \texttt{\href{mailto:jtetreault@dataminr.com}{jtetreault@dataminr.com}},
  \texttt{\href{mailto:marine@cs.umd.edu}{marine@cs.umd.edu}}

 } 

\date{}

\begin{document}
\maketitle
\begin{abstract}
This paper reviews and summarizes human evaluation practices described in $97$ style transfer papers with respect to three main evaluation aspects: style transfer, meaning preservation, and fluency.  In principle, evaluations by human raters should be the most reliable.  However, in style transfer papers, we find that protocols for human evaluations are often underspecified and not standardized, which hampers the reproducibility of research in this field and progress toward better human and automatic evaluation methods.

\end{abstract}

%%%%%% Introduction %%%%%%%%%%%%%%%%%%%%%
\section{Introduction}

Style Transfer (\textsc{st}) in \textsc{nlp} refers to a broad spectrum of text generation tasks that aim to rewrite a sentence to change a specific attribute of language use in context while preserving others (e.g., make an informal request formal, Table~\ref{tab:st_examples}). With the success of deep sequence-to-sequence models and the relative ease of collecting data covering various stylistic attributes,\jt{is this easy?  Seems like the fact there are just a few ST datasets woud mean the opposite? } neural\jt{seems like we can ditch the neural part? ST itself is popular neural or not?} \textsc{st} is a popular generation task with more than $100$ papers published in this area over the last $10$ years. 

Despite the growing interest that \textsc{st} receives from the \textsc{nlp} community, 
progress is hampered by the lack of standardized evaluation practices. 
One practical aspect that contributes to this problem
is the conceptualization and formalization of styles in natural language. 
According to a survey of neural style transfer by \citet{jin2021deep}, in the context of \textsc{nlp}, \textsc{st} is used to refer to tasks where styles follow a linguistically motivated dimension of language variation (e.g., formality), and also to tasks where the distinction between style and content is implicitly defined by data (e.g., positive or negative sentiment). 
\begin{figure}[!ht]
    \centering
    \includegraphics[scale=0.55]{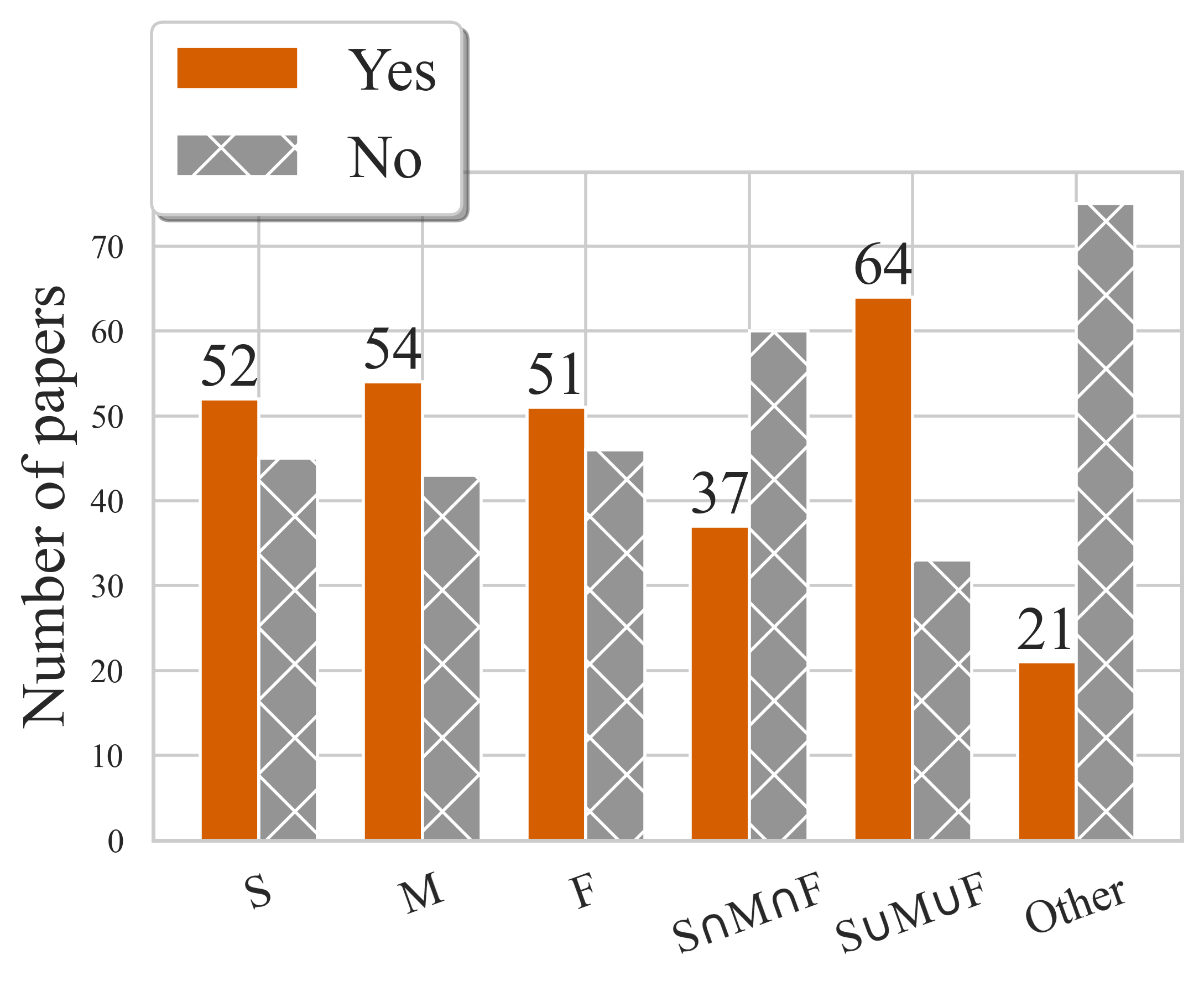}
    \caption{Number of papers employing human evaluations for style transfer (\textsc{s}), meaning preservation (\textsc{m}),
    fluency (\textsc{f}), all of them (\textsc{s}$\cup$\textsc{m}$\cup$\textsc{f}), at least one of them (\textsc{s}$\cap$\textsc{m}$\cap$\textsc{f}),  or another aspect (\textsc{Other}).}
    \label{fig:human_stats}
\end{figure}
\begin{table}[!ht]
    \centering
    \scalebox{0.85}{
    \begin{tabular}{l}
        \rowcolor{gray!10}
        \multicolumn{1}{c}{\textsc{\textbf{formality}}} \\
        Gotta see both sides of the story. \textit{(informal)} \\
        You have to consider both sides of the story. \textit{(formal)}\\
        \addlinespace[1em]
        \rowcolor{gray!10}
        \multicolumn{1}{c}{\textsc{\textbf{sentiment}}} \\
        The screen is just the right size. \textit{(positive)} \\
        The screen is too small. \textit{(negative)} \\
        \addlinespace[1em]
        \rowcolor{gray!10}
        \multicolumn{1}{c}{\textsc{\textbf{author imitation}}} \\     
        Bring her out to me. \textit{(modern)}\\
        Call her forth to me. \textit{(shakespearean)}\\
    \end{tabular}}
    \caption{Examples of three \textsc{st} attributes: formality, sentiment and Shakespearean transfer.}
    \label{tab:st_examples}\vspace{-1.5em}
\end{table}
Across these tasks, \textsc{st} quality is usually evaluated across three dimensions:  style transfer (has the desired attributed been changed as intended?), meaning preservation (are the other attributes preserved?), and fluency (is the output well-formed?) ~\cite{Pang2019UnsupervisedEM, mir-etal-2019-evaluating}. Given the large spectrum of stylistic attributes studied and the lack of naturally occurring references for the associated \textsc{ST} tasks, prior work emphasizes the limitations of automatic evaluation. As a result, progress in this growing field relies heavily on human evaluations to quantify progress among the three evaluation aspects.

Inspired by recent critiques of human evaluations of Natural Language Generation (\textsc{nlg}) systems \citep{howcroft-etal-2020-twenty,lee-2020-annotation,belz-etal-2020-disentangling, belz-etal-2021-systematic,shimorina2021human}, we conduct a structured review of human evaluation for neural style transfer systems as their evaluation is primarily based on human judgments.
Concretely, out of the \textbf{$\mathbf{97}$ papers we reviewed, $\mathbf{69}$ of them resort to human evaluation} ~(Figure~\ref{fig:human_stats}), where it is treated either as a substitute for automatic metrics or as a more reliable evaluation. 

This paper summarizes the findings of the review and raises the following concerns on current human evaluation practices: 
\begin{enumerate}
    \item \textbf{Underspecification} We find that many attributes of the human annotation design (e.g., annotation framework, annotators' details) are  \textit{underspecified} in paper descriptions, which hampers reproducibility and replicability; 
    \item \textbf{Availability \& Reliability} The vast majority of papers do not release the human ratings and do not give details that can help assess their quality (e.g., agreement statistics, quality control), which hurts research on evaluation;
    \item \textbf{Lack of standardization} The annotation protocols are inconsistent across papers which hampers comparisons across systems (e.g., due to possible bias in annotation frameworks).
\end{enumerate}

The paper is organized as follows.  In Section~\ref{sec:procedures}, we describe our procedure for analyzing the $97$ papers and summarizing their evaluations.  In Section~\ref{sec:findings}, we present and analyze our findings.  Finally, in Section~\ref{sec:discussion}, we conclude with a discussion of where the field of style transfer fares with respect to human evaluation today and outline improvements for future work in this area.

%%%%%% Review procedures
\section{Reviewing \textsc{st} Human Evaluation}\label{sec:procedures}

\begin{table*}[!t]
    \centering
    \scalebox{0.89}{
    \begin{tabular}{ll}
    
    \toprule
    \multicolumn{2}{c}{\textbf{\textsc{global criteria}}}\\
    \bottomrule
    \textbf{task(s)} & \textit{\textsc{st} task(s) covered} \\
    \textbf{presence of human annotation} & \textit{presence of human evaluation} \\
    \textbf{annotators' details} & \textit{details on annotator's background/recruitment process} \\
    \textbf{annotators' compensation} & \textit{annotator's payment for annotating each instance} \\
    \textbf{quality control} & \textit{quality control methods followed to ensure reliability of }\\ 
    & \textit{collected judgments}\\ 
    \textbf{annotations' availability} & \textit{availability of collected judgments}\\ 
    \textbf{evaluated systems} & \textit{number of different systems present in human evaluation}  \\
    \textbf{size of evaluated instance set} & \textit{number of instances evaluated for each system}  \\ 
    \textbf{size of annotation set per instance} & \textit{number of collected annotations for each annotated instance}  \\
    \textbf{agreement statistics} & \textit{presence of inter-annotator agreement statistics}  \\
    \textbf{sampling method} &  \textit{method for selecting instances for evaluation from the original test sets} \\
    
    \addlinespace[1.5em]
    
    \toprule
    \multicolumn{2}{c}{\textbf{\textsc{dimension-specific criteria}}}\\
    \bottomrule
    
    \textbf{presence of human evaluation} & \textit{whether there exists human evaluation for a specific aspect} \\ 
    \textbf{quality criterion name} & \textit{quality criterion of evaluated attribute as mentioned in the paper}\\
    \textbf{direct response elicitation} & \textit{presence of direct assessment } \\
    & \textit{(i.e., each instance is evaluated on its own right)}\\
    \textbf{relative judgment type} (if applicable) & \textit{type of relative judgment (e.g., pairwise, ranking, best)}\\
    \textbf{direct rating scale} (if applicable) & \textit{list of possible response values}\\
    \textbf{presence of lineage reference} & \textit{whether the evaluation reuses an evaluation framework from prior work } \\
    \textbf{lineage source} (if applicable) & \textit{citation of prior evaluation framework}\\
    \bottomrule
    \end{tabular}}
    \caption{Descriptions of attributes studied in our structured review.}
    \label{tab:attributes_description}
\end{table*}

\paragraph{Paper Selection} We select papers for this study from the  list compiled by ~\citet{jin2021deep} who conduct a comprehensive review of \textsc{st} that covers the task formulation; evaluation metrics; opinion papers and deep-learning based textual \textsc{st} methods. The paper list contains more than $100$ papers and is publicly available (\url{https://github.com/fuzhenxin/Style-Transfer-in-Text}).
We reviewed all papers in this list to determine whether they conduct either human or automatic evaluation on system outputs for \textsc{st}, and therefore should be included in our structured review.
We did not review papers for text simplification, as it has been studied separately \cite{alva2020data, sikka2020survey} and metrics for automatic evaluation have been widely adopted \cite{xu-etal-2016-optimizing}.
Our final list consists of $97$ papers: $86$ of them are from top-tier \textsc{nlp} and \textsc{ai} venues: \textsc{acl}, \textsc{eacl}, \textsc{emnlp}, \textsc{naacl}, \textsc{tacl}, \textsc{ieee}, \textsc{aaai}, \textsc{n}eur\textsc{ips}, \textsc{icml}, and \textsc{iclr}, and the remaining $11$ are pre-prints which have not been peer-reviewed.

\paragraph{Review Structure} We review each paper based on a predefined set of criteria~(Table~\ref{tab:attributes_description}).
The rationale behind their choice is to collect information on the evaluation aspects that are underspecified in \textsc{nlp} in general as well as those specific to the \textsc{st} task.  For this work, we call the former \textit{global criteria}.
The latter is called \textit{dimension-specific criteria} and is meant to illustrate issues with how each dimension (i.e., style transfer, meaning preservation, and fluency) is evaluated.

\textit{Global criteria} can be split into three categories which describe: (1) the \textsc{st} stylistic attribute, (2) four details about the annotators and their compensation, and (3) four general design choices of the human evaluation that are not tied to a specific evaluation dimension. 

For the \textit{dimension-specific criteria} we repurpose the following \textit{operationalisation} attributes introduced by \citet{howcroft-etal-2020-twenty}: form of response elicitation (direct vs. relative), details on type of collected responses, size/scale of rating instrument, and statistics computed on response values. Finally, we also collect information on the quality criterion for each dimension (i.e., the wording used in the paper to refer to the specific evaluation dimension). 

\paragraph{Process}\jt{can we say that we borrowed from the 2020 INLG paper?}
The review was conducted by the authors of this survey. 
We first went through each of the 97 papers and highlighted the sections which included mentions of human evaluation.\jt{Note that R2 said: "Along the same line, if the paper is accepted, please provide a description of your annotations on the PDFs: what was highlighted? How did you link the highlighted sections to your annotation spreadsheet? Etc."  you may need to include more details}  Next, we developed our criteria by creating a draft based on prior work and issues we had observed in the first step.  We then discussed and refined the criteria after testing it on a subset of the papers.  Once the criteria were finalized, we split the papers evenly between all the authors.
Annotations were spot-checked to resolve uncertainties or concerns that were found in reviewing dimension-specific criteria (e.g., scale of rating instrument is not explicitly defined but inferred from the results discussion) and global criteria (e.g., number of systems not specified but inferred from tables). 
We release the spreadsheet used to conduct the review along with the reviewed \textsc{pdf}s that come with highlights on the human evaluation sections of each paper at \url{https://github.com/Elbria/ST-human-review}.

%%%%%% Findings
\section{Findings}\label{sec:findings}
Based on our review, we first discuss trends of stylistic attributes as discussed in \textsc{st} research through the years ~(\S\ref{sec:stylistic_attributes}), followed by global criteria of human evaluation~(\S\ref{sec:findings_overall}), and then turn to domain-specific criteria~(\S\ref{sec:per_dimension}).

\subsection{Evolution of Stylistic Attributes}\label{sec:stylistic_attributes}

Table~\ref{tab:year_definitions_of_style_stats} presents statistics on the different style attributes considered in \textsc{st} papers since $2011$. First, we observe a significant increase in the number of \textsc{st} papers starting in $2018$ (in $2017$ there were $8$ \textsc{st} papers; the following year there were 28). We believe this can be attributed to the creation of standardized training and evaluation datasets for various \textsc{st} tasks. One example is the Yelp dataset, which consists of positive and negative reviews, and is used for unsupervised sentiment transfer~\cite{yelp}. Another example is the \textsc{gyafc} parallel corpus, consisting of informal-formal pairs that are generated using crowdsourced human rewrites~\cite{rao-tetreault-2018-dear}.
Second, we notice that new stylistic attributes are studied through time (21 over the last ten years), with sentiment and formality transfer being the most frequently studied. 

\begin{table*}[!t]
    \centering
    \scalebox{1.0}{
    \begin{tabular}{lrrrrrrrrrrr}
    \rowcolor{gray!10}
    \textsc{style} & $2011$ & $2012$ & $2016$ & $2017$ & $2018$ & $2019$ & $2020$ & $2021$ & \textsc{total} \\

    \toprule
\rowcolor{blue!10}
            \textit{anonymization}                & & & & &1 & &  &  &1 \\
            \textit{attractiveness}               & & & &1 & & &1 &  &2 \\ \rowcolor{blue!10}
            \textit{author imitation}             & &1 & &2 &2 & 1&5 &  &11 \\
            \textit{debiasing}                    & & & & & & &2 &  &2 \\\rowcolor{blue!10}
            \textit{social register}              & & & & &1 & & & &1 \\
            \textit{expertise}                    & & & & & & &1 &  &1 \\\rowcolor{blue!10}
            \textit{formality}                    & 1 & & & & 1 & 9 & 10 & 3 & 24 \\
            \textit{gender}                       & & & 1 & & 2 & 3 & & & 6 \\\rowcolor{blue!10}
            \textit{political slant}              & & & & & 2 & 1 & 1 & & 4 \\
            \textit{sentiment}                    & & & &4 & 14 & 14 & 18 & 3 & 53\\\rowcolor{blue!10}
            \textit{romantic/humorous}            & & & &  &2 &1 & 1 &  &4 \\
            \textit{simile}                       & & & & & & &1 & &1 \\\rowcolor{blue!10}
            \textit{excitement}                   &  & & & &  &  &1 &   & 1\\
            \textit{profanity}                    & & & & & & &1 &  &1 \\\rowcolor{blue!10}
            \textit{prose}                        & & & & 1&1 & & &  &2 \\
            \textit{offensive language}           & & & & &1 & & 1 &  &2 \\\rowcolor{blue!10}
            \textit{multiple}                     & & & & & &1 &1 &  &2 \\
            \textit{persona}                      & & &1 &  &  &1 &1 &  & 3 \\\rowcolor{blue!10}
            \textit{poeticness}                   & & & &  &  &  &1 &  &1 \\
            \textit{politeness}                   & & & &  &1 &  &1 &  & 2 \\\rowcolor{blue!10}
            \textit{emotion}                   & & & &  & &  & 1 &  &  1\\

            \rowcolor{gray!10}
            \textsc{\textbf{total}}         & $\mathbf{1}$ & $\mathbf{1}$ & $\mathbf{2}$ & $\mathbf{8}$ & $\mathbf{28}$
            & $\mathbf{31}$ & $\mathbf{48}$ & $\mathbf{6}$ & $\mathbf{125}$ \\
            \toprule
        \end{tabular}}
        \caption{Number of \textsc{st} papers per stylistic attribute across years. Some papers evalute multiple style attributes.}
        \label{tab:year_definitions_of_style_stats}\vspace{1em}
\end{table*}

\subsection{Global Criteria}\label{sec:findings_overall}
\paragraph{Annotators} Table~\ref{tab:annotators} summarizes statistics about how papers describe the background of their human judges. The majority of works ($38\%$) rely on crowd workers mostly recruited using the Amazon Mechanical Turk crowdsourcing platform. Interestingly, for a substantial number of evaluations~($45\%$), it is unclear who the annotators are and what their background is. In addition, we find that information about how much participants were compensated is missing from all but two papers.
Finally, many papers collect $3$ independent annotations, although this information is not specified in a significant percentage of evaluations~($42\%$). In short, the ability to replicate a human evaluation from the bulk of current research is extremely challenging, and in many cases impossible, as so much is underspecified. 

\begin{table*}[!ht]
    \centering
    \scalebox{1.0}{
    \begin{tabular}{llr}
    
    \textsc{crowd-sourcing} &  \textsc{paper's description of annotators} & \textsc{count}\\

    \addlinespace[0.3em]
    \hline
    \addlinespace[0.3em]

    \multirow{2}{*}{\textsc{yes}} &  \it  ``qualification test'' & $6$   \\
                                  &  \it ``number of approved \textsc{hit}s' &  $2$ \\
                                  & \it  ``hire Amazon Mechanical Turk workers'' &  $18$ \\
    
    \addlinespace[0.3em]
    \hline
    \addlinespace[0.3em]

    \multirow{5}{*}{\textsc{no}} & \it ``bachelor or higher degree; independent of the authors’  & \multirow{5}{*}{$12$}  \\
                    &             \it``research group'', ``annotators with linguistic background''  &\\
                    &             \it ``well-educated volunteers'', ``graduate students in  & \\
                    &             \it computational linguistics'' ``major in linguistics''  & \\
                    &             \it ``linguistic background'', ``authors'' & \\

    \addlinespace[0.3em]
    \hline
    \addlinespace[0.3em]
    
    \multirow{2}{*}{\textsc{unclear}}   &   \it ``individuals'', ``human judges'', ``human annotators''
    & \multirow{2}{*}{$31$}\\
        &                      \it ``unbiased human judges'', ``independent annotators'' & \\

    \end{tabular}}
    \caption{Annotators' background for human evaluation as described in \textsc{st} papers. 
    }
    \label{tab:annotators}
    %   \vspace{-2em}
\end{table*}
\paragraph{Annotations' Reliability} Only $31\%$ of evaluation methods that rely on crowd-sourcing employ quality control (\textsc{qc})
methods. The most common \textsc{qc} strategies are to require workers to pass a qualification test~\cite{Jinetal19, li-etal-2016-persona, ma-etal-2020-powertransformer, Pryzant_Diehl_Martinez_Dass_Kurohashi_Jurafsky_Yang_2020} to hire the top-ranked workers based on pre-computed scores that reflect the number of their past approved tasks~\cite{krishna-etal-2020-reformulating, li-etal-2019-domain}, to use location restrictions~\cite{krishna-etal-2020-reformulating},
or to perform manual checks on the collected annotations~\cite{rao-tetreault-2018-dear,briakou2021xformal}. Furthermore, only $20\%$ of the papers report inter-annotator agreement statistics, and only $4$ papers release the actual annotations to facilitate the reproducibility and further analysis of their results.  Without this information, it is difficult to replicate the evaluation and compare different evaluation approaches.\jt{can we say something about how many papers actually release the templates? - I think it's just the XFORMAL paper right?} \eb{I am not 100 percent sure about this one}

\paragraph{Data Selection} Human evaluation is typically performed on a sample of the test set used for automatic evaluation.
Most works~($62\%$) sample instances randomly from the entire set, with a few exceptions that employ stratified sampling according to the number of stylistic categories considered (e.g., random sampling from positive and negative classes for a binary definition of style).
For $25\%$ of \textsc{st} papers information on the sampling method is not available. 
Furthermore, the sample size of instances evaluated per system varies from $50$ to $1000$, with most of them concentrated around $100$. 

\subsection{Dimension-specific Criteria}\label{sec:per_dimension}
\paragraph{Quality Criterion Names} Table~\ref{tab:quality_criterion_names} summarizes the terms used to refer to 
the three main dimensions of style transfer, meaning preservation, and fluency. As \citet{howcroft-etal-2020-twenty} found in the context of \textsc{nlg} evaluation, we see that the names of these dimensions are not standardized for the three \textsc{st} evaluation dimensions. Each dimension has at least six different ways that past literature has referred to them.  We should note that even with the same name, the nature of the evaluation is not necessarily the same across \textsc{ST} tasks: for instance, what constitutes content preservation differs in formality transfer and in sentiment transfer, since the latter arguably changes the semantics of the original text. While fluency is the aspect of evaluation that might be most generalizable across \textsc{ST} tasks, it is referred to in inconsistent ways across papers which could lead to different interpretations by annotators.  For instance, the same text could be rated as natural but not grammatical. Overall, the variability in terminology makes it harder to understand exactly what is being evaluated and to compare evaluation methods across papers. 

\paragraph{Rating Type} Table~\ref{tab:eval_frameword_across_years} presents statistics on the rating type (direct vs. relative) per dimension over time. \textit{Direct rating} refers to evaluations where each system output is assessed in isolation for that dimension.  \textit{Relative rating} refers to evaluations where two or more system outputs are compared against each other.
Rating types were more inconsistently used before $2020$, with recent convergences toward direct assessment. %ranking. 
Among papers that report rating type, direct assessment is the most frequent approach for all evaluation aspects over the years $2018$ to $2021$.

\begin{table}[!t]
    \centering
    \scalebox{1.0}{
    \begin{tabular}{  p{7.5cm}  }
    
    % Style
    \rowcolor{gray!10}
    \textbf{\textsc{style}}   \\
    \it attribute compatibility, 
    formality, 
    politeness level, 
    sentiment, 
    style transfer intensity,
    attractive captions, 
    attribute change correctness,
    bias, 
    creativity, 
    highest agency,
    opposite sentiment,
    sentiment,
    sentiment strength,
    similarity to the target attribute, 
    style correctness,
    style transfer accuracy,
    style transfer strength, 
    stylistic similarity,  
    target attribute match,
    transformed sentiment degree. \\
    
    \addlinespace[1.5em]
    \rowcolor{gray!10}
    \textbf{\textsc{meaning}}   \\   
    \it content preservation, % 21
    meaning preservation, %3
    semantic intent,   %1
    semantic similarity, %3
    closer in meaning to the original sentence, %1
    content preservation degree, %1                 
    content retainment, %1
    content similarity, %1
    relevance, %1
    semantic adequacy. % 1
    \\
     
     \addlinespace[1.5em]
    \rowcolor{gray!10}
    \textbf{\textsc{fluency}}   \\    
    \it
    fluency,                                      %23
    grammaticality,                               % 4
    naturalness,                                  % 1
    gibberish language,   %  1
    language quality.                             % 1
\\
    \end{tabular}
    }
    \caption{Quality criterion names used in \textsc{st} human evaluation descriptions for the three evaluation dimensions.}\label{tab:quality_criterion_names}\vspace{1em}
\end{table}
\begin{table}[!h]
    \centering
    \scalebox{0.88}{
\begin{tabular}{lrrrrrrrrrrr@{\hskip 3.8in}r}
     & \rot{$2011$} & \rot{$2012$} & \rot{$2016$} & \rot{$2017$} & \rot{$2018$} & \rot{$2019$} & \rot{$2020$} & \rot{$2021$} &  \rot{Total}\\

    \rowcolor{gray!10}
    \multicolumn{10}{c}{\textbf{\textsc{style}}}\\
    
    \addlinespace[0.5em]
    
    \textsc{direct} & 1 &1 & & 1&8 &10 &12 &4 & 40 \\
    \textsc{relative} & & &1 & &4 &7 & & & 12 \\
    \textsc{none} & &  &2 &6 &11 &11 &15 & & 45 \\
    
    \addlinespace[0.5em]
    
    \rowcolor{gray!10}
    \multicolumn{10}{c}{\textbf{\textsc{meaning}}} \\
    
    \addlinespace[0.5em]
    
    \textsc{direct} & &1 & & &12 &10 &18 &4 & 45\\
    \textsc{relative} & & & &1 & &4 &4 & & 9\\
    \textsc{none} & 1 & &2 &7 &8 &11 &14 & & 43\\
    
    \addlinespace[0.5em]
    
    \rowcolor{gray!10}
    \multicolumn{10}{c}{\textbf{\textsc{fluency}}} \\
    
    \addlinespace[0.5em]
    
    \textsc{direct} & &1 & &1 &10 &10 &19 &4 & 45\\
    \textsc{relative} &  & & & &4 &2 & & & 6\\
    \textsc{none} & 1 & &1 &2 &8 &6 &7 & & 46\\

    \end{tabular}}
    \caption{Number of papers using each rating type for the three evaluation dimensions across years.} 
    \label{tab:eval_frameword_across_years}
    %  \vspace{-1em}
\end{table}

\paragraph{Possible Responses} Tables~\ref{tab:style_tab}, \ref{tab:mean_tab}, and \ref{tab:flu_tab}
summarize the range of responses elicited for direct and relative ratings. 
They cover diverse definitions of scales within each rating type. 
Across evaluation aspects, the dominant evaluation framework is \textbf{direct ratings} on a \textbf{5-point scale}.  However, while that configuration is what the field tends to focus on, there is clearly a wide array of choices that the field also considers which, once again, makes comparing human evaluations head to head very difficult.

\begin{table*}
    \raggedright
    {\large  %
    \scalebox{0.8}{
    \begin{tabular}{lcrrl}
    
  \multirow{14}{*}{\textsc{Direct}}     & \multirow{14}{*}{(40)} \rdelim\{{14}{*}[]   & 
  Rating Scale
  & (1) & [-2,-1,0,1,2]  \\
                 &     &          & (3) & [-3,-2, -1, 0, 1, 2, 3] \\
            &    &  &       (1) & [polite, slightly polite, neutral, slightly rude, rude] \\
                 &     &          & (4) & [positive, negative, neutral] \\
                 &     &          & (1) & [positive, negative, relaxed, annoyed] \\
                  &     &          & (1) & [more formal, more informal, neither] \\

                 &     &    & (2)   & [0,1,2]  \\
                 &     &          & (2)   & [1,2,3]  \\
                  &     &          & (1)   & [0,1,2,3,4,5] \\
                 &     &          & \cellcolor{blue!25}(19)  & \cellcolor{blue!25}[1, 2, 3, 4, 5] \\
                 &     &          & (2)   & [1,2,3,4,5,6,7,8,9,10] \\
                 &     &          & (1)   & binary \\

                  \addlinespace[0.3cm]
                 &     &     Not available      & (2)   & \\   
                     \addlinespace[0.3cm]

    \multirow{2}{*}{\textsc{Relative}} & \multirow{2}{*}{(12)} \rdelim\{{2}{*}[]  & Best selection & (5) &   \\
                 &  & Pairwise        & (7) & \\
    \end{tabular}}\caption{\textbf{Style} results. Numbers in parentheses denote paper counts per category. The most popular rating type across each dimension is highlighted.}\label{tab:style_tab}  \vspace{4em}
    \scalebox{0.8}{
    \noindent
    \begin{tabular}{lcrrl}

    \multirow{10}{*}{\textsc{Direct}}     & \multirow{10}{*}{(45)} \rdelim\{{10}{*}[]   & Rating Scale
    & (1) & [-2,-1,0,1,2]  \\

                 &     &   & (6)   & [0,1,2]  \\
                 &     &    & (1)   & [1,2,3]  \\
                   &     &    & (1)   & [1,2,3,4]  \\
                 &     &          & \cellcolor{blue!25}(25)  & \cellcolor{blue!25}[1, 2, 3, 4, 5] \hspace{15cm}\\
                  &     &          & (1)   & [0,1,2,3,4,5] \\
                 &     &          & (4)   & [1,2,3,4,5,6] \\
                 &     &          & (3)   & [1,2,3,4,5,6,7,8,9,10] \\
                
                    \addlinespace[0.3cm]
                 &     &     Not available      & (3)   & \\   
                     \addlinespace[0.3cm]

    \multirow{3}{*}{\textsc{Relative}} & \multirow{3}{*}{(9)} \rdelim\{{3}{*}[]  & Best selection & (3) &   \\
                 &  & Pairwise        & (3) & \\
                 &  & Ranking        & (3) & \\
    
    \end{tabular}}\caption{\textbf{Meaning Preservation} results. Numbers in parentheses denote paper counts per category. The most popular rating type across each dimension is highlighted.}\label{tab:mean_tab} \vspace{4em}     

    \scalebox{0.8}{
    \begin{tabular}{lcrrl}
    \multirow{13}{*}{\textsc{Direct}}     & \multirow{13}{*}{(45)} \rdelim\{{13}{*}[]   & 
    Rating Scale
    & (1) & ["easy to understand", "some grammar errors", "impossible to understand"] \\
    &  & & (1) &  ["incorrect", "partly correct", "correct"]  \\

                 &     &   & (1)   & [0,1]  \\
                 &     &          & (3)   & [0,1,2]  \\
                 &     &          & (2)   & [1,2,3]  \\
                 &     &          & (4)    & [1,2,3,4]\\
                 &     &          & (1)    & [0,1,2,3,4]\\
                 &     &          & \cellcolor{blue!25}(26)  & \cellcolor{blue!25}[1, 2, 3, 4, 5] \\
                  &     &          & (1)    & [0,1,2,3,4,5]\\
                  &     &          & (1)    & [1,2,3,4,5,6]\\
                 &     &          & (2)   & [1,2,3,4,5,6,7,8,9,10] \\
         
                    \addlinespace[0.3cm]
                 &     &     Not available      & (2)   & \\   
                     \addlinespace[0.3cm]

    \multirow{3}{*}{\textsc{Relative}} & \multirow{3}{*}{(6)} \rdelim\{{3}{*}[]  & Best selection & (1) &   \\
                 &  & Pairwise        & (4) & \\
                 &  & Ranking        & (1) & \\
                 
    \end{tabular}}    
    \caption{\textbf{Fluency} results. Numbers in parentheses denote paper counts per category. The most popular rating type across each dimension is highlighted.}\label{tab:flu_tab}  
    }
    \label{tab:rating_types}
\end{table*}

\begin{figure*}[!t] 
    \centering
    \includegraphics[width=0.85\linewidth]{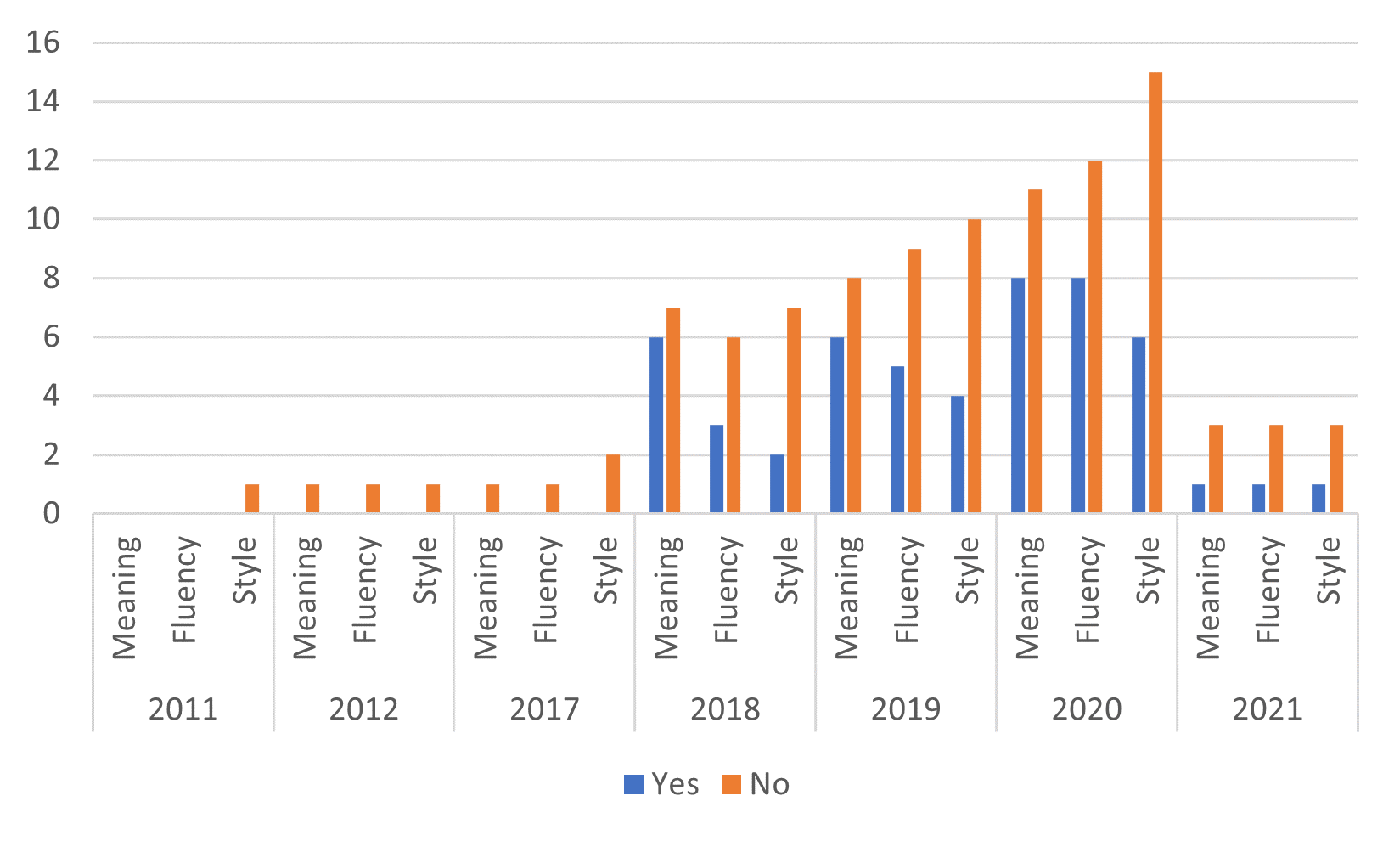}
    \caption{Lineage statistics (i.e., number of papers) for each \textsc{st} evaluation aspect over years.} \label{fig:lineage}
\end{figure*}

\paragraph{Lineage}
Figure~\ref{fig:lineage}\jt{y axis needs a caption (number of papers)} shows how often the human evaluation setup used in each reviewed paper is based on cited prior work, for each dimension over time. Only $19\%$ of papers repurpose or reuse some prior work for the evaluation of style. Most of these papers target \textsc{st} for formality or sentiment.
Even when evaluating fluency or meaning preservation, more than $50\%$ of the papers\jt{is it 50\% of the 19\% ?} do not refer to any prior work. This is striking because it suggests that there is currently not a strong effort to replicate prior human evaluations.

\jt{looking at Figure 2 and setting aside 2021, could we say that as time goes on, a >smaller< percentage of work is reusing/repurposing prior work, at least explicitly in the form of a citation.  This is a bit concerning as it could mean that newer papers are increasingly doing evaluations in new ways, further making head to head comparison with prior work impossible - in particular for style}

\jt{just noticed this with Figure 2 - the order should be style, meaning, fluency as we have it in the text}

For papers that mention lineage, the most common-set up for evaluating meaning preservation ($24\%$) and fluency ($28\%$) is \newcite{li-etal-2018-delete}. $43\%$ of \textsc{st} papers that work on sentiment also refer to \newcite{li-etal-2018-delete}. Some papers follow  \newcite{agirre-etal-2016-semeval} for measuring textual similarity, \newcite{heilman-etal-2014-predicting} for grammaticality and \newcite{pavlick2016empirical} for formality. 

%%%%%%% Discussion
\section{Discussion \& Recommendations}\label{sec:discussion}

\subsection{Describing Evaluation Protocols}

Our structured review shows that human evaluation protocols for \textsc{ST} are mostly underspecified and lack standardization, which fundamentally hinders progress, as it is for other \textsc{NLG} tasks \citep{howcroft-etal-2020-twenty}. The following attributes  are commonly underspecified:
\begin{enumerate}
\item details on the procedures followed for recruiting annotators (i.e., linguistic background of expert annotators or quality control method employed when recruiting crowd-workers) 
\item annotator's compensation to better understand their motivation for participating in the task,
\item inter-annotator agreement statistics,
\item number of annotations per instance ($3$-$5$ is the most popular choice of prior work),
\item number of systems evaluated,
\item number of instances annotated (minimum of $100$ based on prior works),
\item selection method of the annotated instances (suggestion is same random sampled for all annotated systems).
\item detailed description of evaluated frameworks per evaluation aspect (e.g., rating type, response of elicitation). 
\end{enumerate}

Furthermore, we observe that annotated judgments are hardly ever made publicly available and that, when specified, evaluation frameworks are not standardized.

As a result, our first recommendation is simply to include all these details when describing a protocol for human evaluation of \textsc{st}. We discuss further recommendations next.

\subsection{Releasing Annotations}

Making human-annotated judgments available would enable the development of better automatic metrics for  \textsc{st}. If all annotations had been released with the papers reviewed, we estimate that more than $10$K human judgments per evaluation aspect would be available. Today this would suffice to train and evaluate dedicated evaluation models.

In addition, raw annotations can shed light on the difficulty of the task and nature of the data: they can be aggregated in multiple ways \citep{oortwijn-etal-2021-interrater}, or used to account for annotator bias in model training~\cite{beigman-beigman-klebanov-2009-learning}. Finally, releasing annotated judgments makes it possible to replicate and further analyze the evaluation outcome~\cite{belz-etal-2021-systematic}.

\subsection{Standardizing Evaluation Protocols}

Standardizing evaluation protocols is
key to establishing fair comparisons across systems~\cite{belz-etal-2020-disentangling} and to improving evaluation itself.

Our survey sheds light on the most frequently used \textsc{st} frameworks in prior work. Yet more research is needed to clarify how to evaluate, compare and replicate the protocols. For instance, 
\citet{mir-etal-2019-evaluating} point to evidence that relative judgments can be more reliable than absolute judgments \cite{StewartBrownChater2005},
as part of their work on designing automatic metrics for \textsc{st} evaluation. However, research on human evaluation of machine translation shows that this can change depending on the specifics of the annotation task:  relative judgments were replaced by direct assessment when \citet{GrahamBaldwinMoffatZobel2013} showed that both intra and inter-annotator agreement could be improved by using a continuous rating scale instead of the previously common five or seven-point interval scale \cite{Callison-BurchFordyceKoehnMonzSchroeder2007}.

For \textsc{st}, the lack of detail and clarity in describing evaluation protocols makes it difficult to improve them, as has been pointed out for other \textsc{nlg} tasks by \citet{shimorina2021human} who propose evaluation datasheets for clear documentation of human evaluations, \citet{lee-2020-annotation} and \citet{van2020human} who propose best practices guidelines, and \citet{belz-etal-2020-disentangling, belz-etal-2021-systematic} who raise concerns regarding reproducibility. This issue is particularly salient for \textsc{st} tasks where stylistic changes are defined implicitly by data \citep{jin2021deep} and where the instructions given to human judges for style transfer might be the only explicit characterization of the style dimension targeted. Furthermore, since \textsc{st} includes rewriting text according to pragmatic aspects of language use, who the human judgments are matters since differences in communication norms and expectations might result in different judgments for the same text.

Standardizing and describing protocols is also key to assessing the alignment of the evaluation with the models and task proposed \citep{hamalainen-alnajjar-2021-great}, and to understand potential biases and ethical issues that might arise from, e.g., compensation mechanisms \citep{vaughan,schoch-etal-2020-problem,shmueli2021beyond}.

\bibliographystyle{acl_natbib}
\bibliography{anthology,acl2021,gem}

%\appendix

\end{document}